\documentclass[letterpaper, 10 pt, conference]{ieeeconf}  

\IEEEoverridecommandlockouts                              
\usepackage{amsmath}
\usepackage{graphicx}      

\title{\LARGE \bf
Design and Control of an Aerial Manipulator for Contact--based Inspection
}

\author{Varun Nayak$^{1}$, Christos Papachristos$^{2}$, and Kostas Alexis$^{3}$
\thanks{$^{1}$Varun Nayak is an undergraduate student pursuing mechanical engineering at Birla Institute of Technology and Science (BITS), Pilani, Goa, India
        {\tt\small f20140086@goa.bits-pilani.ac.in}}%
\thanks{$^{2}$ Christos Papachristos, and Kostas Alexis are with the Autonomous Robots Lab, Department of Computer Science and Engineering, University of Nevada, Reno
        {\tt\small cpapachristos@unr.edu, kalexis@unr.edu}}%
}

\begin{document}

\maketitle
\thispagestyle{empty}
\pagestyle{empty}

\begin{abstract}

 Manipulator dynamics, external forces and moments raise issues in stability and efficient control during aerial manipulation. Additionally, multirotor Micro Aerial Vehicles impose stringent limits on payload, actuation and system states. In view of these challenges, this work addressed the design and control of a 3-DoF serial aerial manipulator for contact inspection. A lightweight design with sufficient dexterous workspace for NDT (Non-Destructive Testing) inspection is presented. This operation requires the regulation of normal force on the inspected point. Contact dynamics have been discussed along with a simulation of the closed-loop dynamics during contact. The simulated controller preserves inherent system nonlinearities and uses a passivity approach to ensure the convergence of error to zero. A transition scheme from free-flight to contact was developed along with the hardware and software frameworks for implementation. This paper concludes with important drawbacks and prospects.

\end{abstract}

\section{INTRODUCTION}

Unmanned Aerial Vehicles (UAV) are being widely used today for applications such as mapping, inspection, exploration and photography~\cite{bircher2016receding,NBVP_ICRA_16,SIP_AURO_2015,ADBS_AURO_2015,BABOOMS_ICRA_15}. An important observation regarding most applications of aerial robots is that they do not involve any kind of contact with the environment. This phenomenon can be attributed to several challenges. Traditional robotic manipulators whose base links are fixed to the ground (earth) are able to efficiently dissipate external and inertial forces encountered during manipulation. The reaction forces necessary for balancing external as well as inertial forces acting on the manipulator are available immediately owing to the fixed contact relationship of the base link with the infinitely dissipative ground. However, this is not the case for aerial manipulators as multirotor vehicle dynamics are relatively sensitive and ``slow'' because of inherent aerodynamic forces as well as inertia. Therefore, the task of maintaining the desired vehicle attitude against these external contact forces becomes a challenge. Since multirotors are unstable in general, compensator design and stability for closed-loop contact dynamics must be addressed. Along with performing control during contact, the transition between the free-flight regime and the contact regime can present some complications. Physical properties of the UAV platform such as payload, limits on thrust and inertia lead to intricate design challenges. 

\begin{figure}[t]
	\centering
		\includegraphics[width=0.85\linewidth, height=5.5cm]{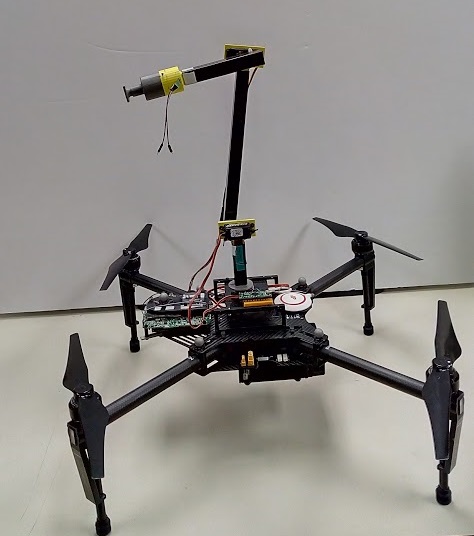}
		\caption[Photograph of the Fully Assembled Aerial Manipulator System Developed]{The fully assembled aerial manipulator system developed in this work. The image shows the 3-DoF RRR serial manipulator mounted on top of a DJI Matrice 100 quadrotor.}
	\label{fig:a}
\end{figure} 

Researchers have employed a variety of approaches in design and control methodologies in aerial manipulation applications~\cite{bellicoso2015design,fumagalli2012modeling,heredia2014control,kamel2016design,kamel2016full,keemink2012mechanical,murray2017mathematical,ortega1997passivity,pounds2011uav,sa2017dynamic,scholten2013interaction,wopereis2016mechanism,papachristos2014efficient}. Previous research touches on topics such as kinematics for workspace and dexterity, full-body control, lightweight design, accurate end-effector position control, interaction control, etc. The work in~\cite{kamel2016design} developed a novel parallel manipulator with a large workspace and current-based torque control to employ impedance control schemes. The work in~\cite{bellicoso2015design} designed a lightweight 5-DoF aerial manipulator for pick and place applications. It has a smart self-folding mechanism to minimize space occupation and static CoG imbalance. A special differential mechanism to cancel attitude disturbances was also designed. The research presented in~\cite{keemink2012mechanical} used delta-kinematics for designing a fast and precise aerial manipulator for contact-based inspection. The sophisticated delta structure possesses compliance allowing for slight tracking errors. The researchers in~\cite{wopereis2016mechanism} demonstrated the design and operation of a unique superstructure manipulator that has the ability to perch on a vertical surface through impact. It is lightweight and possesses unilateral compliance for perch and release operations.

The authors in~\cite{heredia2014control} developed a 7-DoF aerial manipulator for heavy payloads. Hence, it used a backstepping-based controller for the multi-rotor, which considers full coupled dynamics and the rapidly shifting CoG. An admittance-based manipulator controller is outlined in the paper. The contribution in~\cite{kamel2016full} presented a multi-objective full-body controller for the system described in. The fast dynamics of the parallel manipulator ensured efficient kinematic tracking. Furthermore, the work in~\cite{scholten2013interaction} addressed the problem of interaction in order to track desired contact force using hybrid control for the manipulator,  using an impedance-based controller for position and a PI controller for regulating normal force. The authors in~\cite{fumagalli2012modeling} presented the design and control of a parallel aerial manipulator for industrial inspection. The approach considered the environment as a compliant contact and used the Hunt-Crossley interaction model.

This work modeled the NDT (Non-Destructive Testing) inspection task as a force regulation problem~\cite{fumagalli2012modeling} and designed a lightweight 3-DoF RRR manipulator with sufficient dexterity. A planar model of the dynamics during contact was developed, along with a passivity-based PD controller while preserving the nonlinearity of the model. Keeping in mind actuation limits and stability requirements, a simulation of the closed-loop dynamics is presented with a smooth free-flight to contact transition scheme. This paper also describes the hardware and software framework for implementation.

\section{MANIPULATOR DESIGN}

Although most aerial manipulators have been mounted below the plane of the rotor and t were designed for pick and place tasks. This makes sense spatially as well as for stability. However, since our objective was to achieve continuous contact in the horizontal plane, the manipulator was mounted as a super-structure. Placing the manipulator on top makes the external force a stabilizing moment and also reduces the required manipulator workspace.

\begin{figure}[ht]
	\centering
		\includegraphics[width=0.8\linewidth, height=4.7cm]{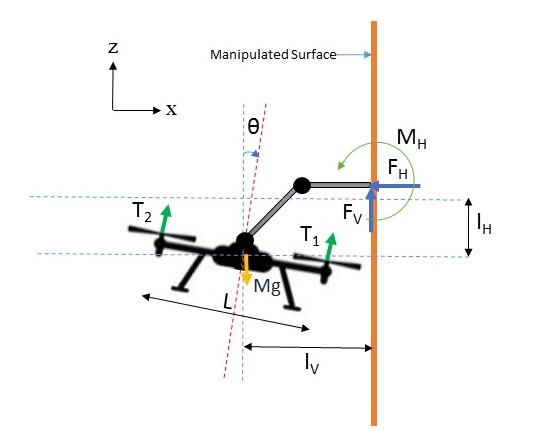}
		\caption[Free Body Diagram for Static Analysis]{A 2-D free body diagram of the aerial manipulator in contact with the manipulated surface. The figure shows all considered forces, moments and the coordinate system used in the analysis. $\theta$ is the pitch angle, $F_H$ is the horizontal contact force, $F_V$ is the vertical contact force, $T_1$ and $T_2$ are the thrust values as shown in the figure and $L$, $l_v$ and $l_H$ are the geometric parameters as shown.}
	\label{fig:fbd}
\end{figure}

\begin{figure}[ht]
	\centering
		\includegraphics[width=0.9\linewidth, height=5cm]{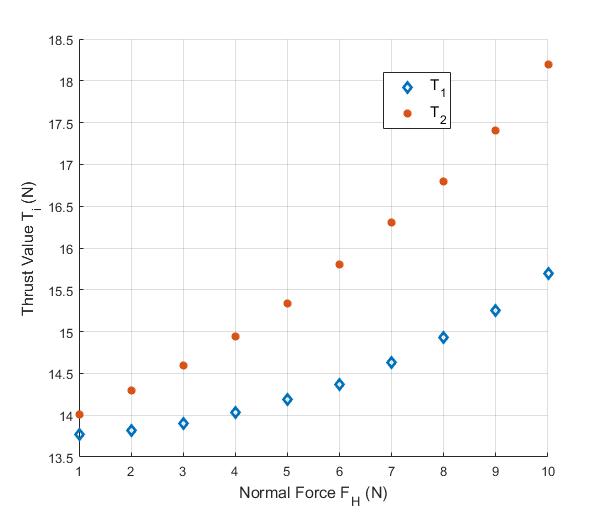}
		\caption[Static Simulation Results of Contact]{Static simulation results during contact: The plot shows the variation of the individual thrust components with respect to the contact force required. A nominal friction force $\mu = 0.3$ is taken in the downward direction (conservative) has been considered. The nonlinearity for large values is evident.}
	\label{fig:staticsim}
\end{figure}

Additionally, a static analysis of the system in contact showed that system nonlinearities are more pronounced when the contact force exceeds 6\,N. Hence, it was decided to preserve this inherent nonlinearity for the analysis.

\begin{figure}[htbp]
	\centering
		\includegraphics[width=0.8\linewidth, height=4.5cm]{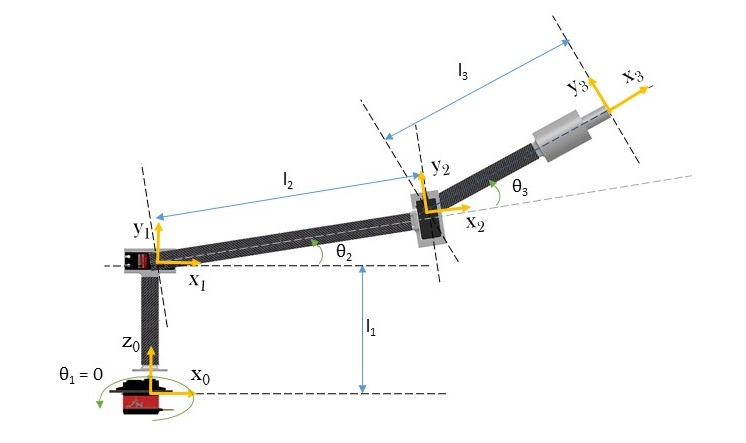}
		\caption[Side-View of RRR Manipulator Assembly]{Side-view of RRR manipulator assembly showing the coordinate systems, joint angles and other definitive geometric parameters}
	\label{fig:sidev}
\end{figure}

\subsection{Kinematics}

The manipulator is an RRR type with the base joint for yaw decoupling of the arm from the multirotor heading. This allows for better error correction during manipulator position tracking. This makes the rest of the arm a planar 2-DoF RR type, which provides sufficient dexterity without being too heavy.

The forward kinematics of the manipulator were developed using the Denavit-Hartenberg (DH) convention. The DH parameters of the manipulator are shown in TABLE I. The inverse kinematics were developed analytically. The first joint angle is obtained using spherical coordinate decoupling. Following this, solving for the inverse kinematics of a 2-DoF RR manipulator is trivial.  

\begin{table}[h]
\begin{center}
\begin{tabular}{ |c||c|c|c|c| } 
 \hline
 Transform from link $i-1$ to $i$ & $r_i$ & $d_i$ & $\alpha_i$ & $\theta_i$ \\ \hline
 $i=1$ & 0 & $l_1$ & $\pi/2$ & $\theta_1$  \\  \hline
 $i=2$ & $l_2$ & 0 & 0 & $\theta_2$ \\  \hline
   $i=3$ & $l_3$ & 0 & 0 & $\theta_3$ \\  \hline
  \end{tabular}
 \caption[Denavit Hartenberg Parameters of the RRR Serial Manipulator]{DH Parameters of the RRR Serial Manipulator.}
\end{center}
\end{table}

\begin{figure}[h]
	\centering
		\includegraphics[width=0.9\linewidth, height=5.4cm]{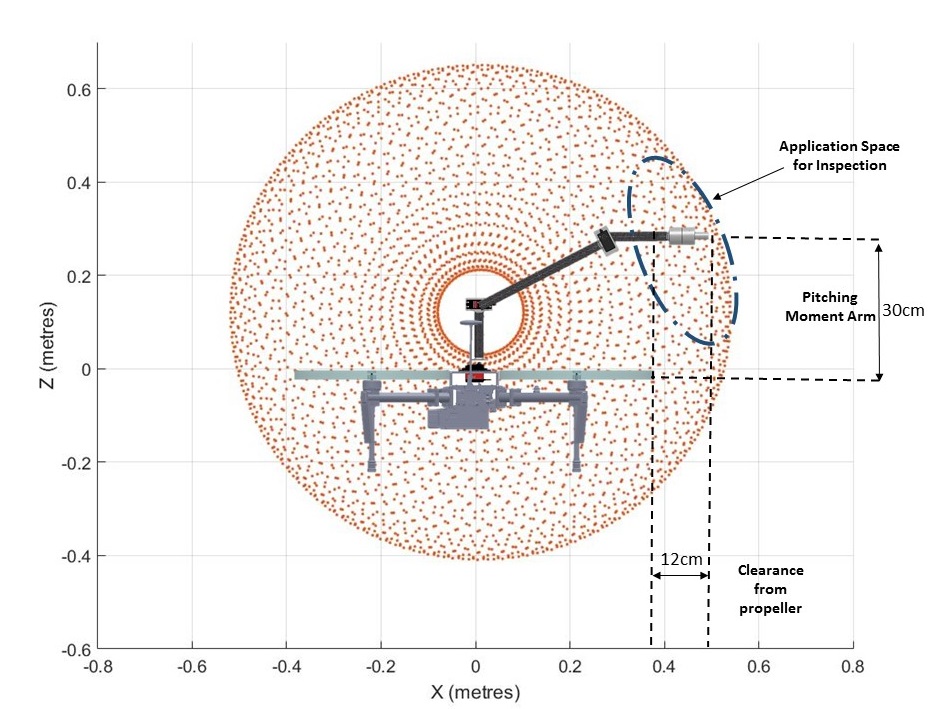}
		\caption[Workspace in X-Z Plane]{The figure shows the manipulator in a typical pose during contact. The true workspace is the displayed workspace revolve-swept about the base joint axis. The application space is the set of points that the end-effector can reach with the desired orientation while satisfying all spatial constraints.}
	\label{fig:wspace}
\end{figure}

In the workspace analysis, the joint torque requirements during manipulation were considered since these needed to be minimized to reduce motor size. The relationship $\boldsymbol{\tau} =  \textbf{J}^T \textbf{F}$ was used. Here $\textbf{F}$ is the end-effector force, $\tau$ is the joint torque vector and $\textbf{J}$ is the manipulator Jacobian.

A typical load-configuration combination for this system is $F_H=6\,N$ with in-plane frictional forces equal to about $3.5\,N$ at the configuration $\theta_1=0^\circ$, $\theta_2=35^\circ$ and $\theta_3= -15^\circ$. The required torques for this configuration from equation (4.14) are computed as $M_1= 1.6\,N.m$, $M_2= 3.1 \,N.m$ and $M_3=1.1\,N.m$. The motors were selected by simulating similar such cases.

\subsection{Compliant Mechanism}

Integration of a compliant mechanism at the end-effector was conceived in order to satisfy the following objectives:
\begin{itemize}
    \item To absorb impact when the quadrotor first comes in contact to slow down the resulting contact dynamics.
    \item To allow for slight curvatures on the surface and small errors in the end-effector pose.
    \item To enable compliant force-sensing at the end-effector.
\end{itemize}

\begin{figure}[ht]
	\centering
		\includegraphics[width=0.8\linewidth, height=4.5cm]{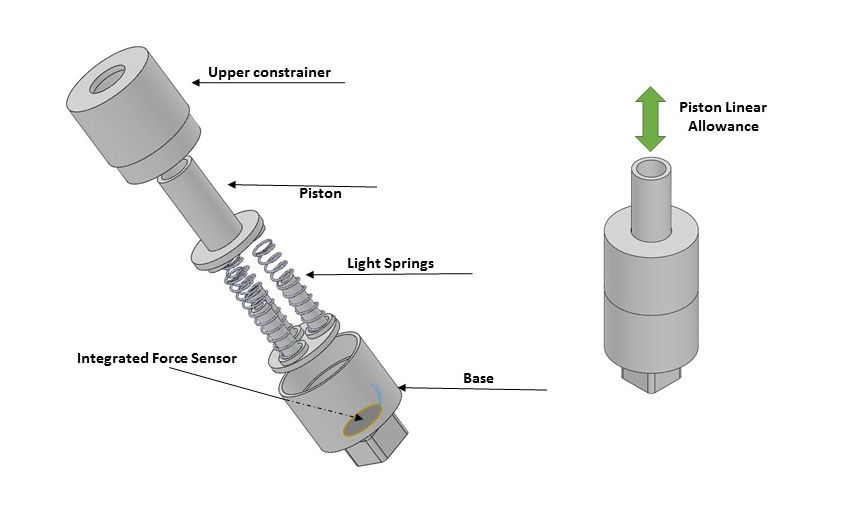}
		\caption[Exploded View of Compliant Mechanism]{Exploded view of the compliant mechanism. The springs are preloaded and force feedback is obtained from a resistive force sensor. The three springs in parallel allow for the contact plane adjusting to surface curvatures and eccentric forces.}
	\label{fig:updowncomp}
\end{figure}

The entire manipulator along with the motors weighed only 327 grams, which was about half of the maximum payload of the quadrotor used for implementation.

\section{CONTACT DYNAMICS AND CONTROL}

The manipulator designed in this work weighed only 327\,grams against the platform weight of 2400\,g. In addition to this, the maximum shift in CoG for the entire configuration space of the aerial manipulator system was only 2\,cm from the vertical axis passing through the centroid of the multirotor. Therefore, no special consideration was given to the effect of the inertial effects of the manipulator with respect to the aerial vehicle in the closed-loop dynamics simulations. However, contact introduces an additional external force on the system. The constraints on the thrust values, thrust rates and state variables during flight necessitates a fundamental closed-loop dynamics analysis. An important assumption was that the nature of the equation of motion in the vertical direction was considered as static. As contact dynamics are fast compared to that during free-flight, the analysis of the controller performance for this direction is crucial. Hence, this assumption was acceptable. The analysis is performed in a 2-D plane and friction has been ignored, compensated by a safety factor.

Applying Newton's 2$^{nd}$ Law, we can write the equations of motion in state-space form. $I_{yy}$ is the moment of inertia about the axis parallel to the y-axis and passing through the CoM of the entire system. For simplicity, the explicit time dependency symbol is dropped hereafter.

\begin{equation}
\begin{bmatrix} \dot{x} \\ \ddot{x} \\ \dot{\theta} \\\ddot{\theta} \end{bmatrix}  =  \begin{bmatrix} \dot{x} \\ -\frac{F_H}{M} \\ \dot{\theta} \\ \frac{F_H l_H}{I_{yy}} \end{bmatrix} + \begin{bmatrix} 0 & 0 \\ \frac{\sin\theta}{M} & \frac{\sin\theta}{M} \\ 0 & 0 \\ -\frac{L}{2I_{yy}} & \frac{L}{2I_{yy}} \end{bmatrix} \begin{bmatrix} T_1 \\ T_2 \end{bmatrix} \\
\end{equation}

This equation is in the vector-valued nonlinear affine (affine in \textbf{u}) form.
$$
\dot{\textbf{x}} = \textbf{f}(\textbf{x}) + \textbf{g}(\textbf{x})\textbf{u}
$$

Here, $\textbf{x}$ is the state vector and $\textbf{u}$ is the control input.

$$ 
\textbf{x} = \begin{bmatrix} x \\ \dot{x} \\ \theta \\\dot{\theta} \end{bmatrix} \hspace{0.2cm}, \hspace{0.2cm} \textbf{u} = \begin{bmatrix} T_1 \\ T_2 \end{bmatrix}
$$. 

The expression for force from the dynamics can be written as follows:

\begin{center}
\begin{eqnarray}
F_H & = & Mg\,\tan\theta - M\ddot{x}
\end{eqnarray}
\end{center}

Since the objective was to regulate the contact force, a general approach would have demanded force feedback control from the end-effector. It is indicative from the equation that the regulation of the pitch angle and the x-position implies the regulation of the normal force. Since force-feedback control (impedance or compliant control) was beyond the scope of this work, this model was used. Thus, the aim was to control the x-position and the pitch angle. The output state to be controlled is expressed by the function $\textbf{y} = h(\textbf{x})$. A passivization approach through a nonlinear PD controller was used to model the control law for this system. As the system is affine in $\textbf{u}$, the control law was derived using the input-output linearization technique.

Insisting that the error convergence should follow second-order dynamics, we can write:

\begin{equation}
\ddot{\textbf{h}}-\ddot{\textbf{h}}_{desired}+ \textbf{k}_2(\dot{\textbf{h}}-\dot{\textbf{h}}_{desired}) +  \textbf{k}_1(\textbf{h}-\textbf{h}_{desired})  =  0
\end{equation}

In the above equations, $\textbf{k}_1$ and $\textbf{k}_2$ are positive definite matrices containing the gains required for stable dynamics. Using Lie derivatives, the control law was derived.

\begin{eqnarray}
\textbf{u} &=&  L_{\textbf{g},\textbf{f}}\textbf{h}^{-1}(- L_{\textbf{f },\textbf{f}}\textbf{h} + \textbf{v})
\end{eqnarray}

The expression for $\textbf{v}$ comes from the second order error dynamics.

\begin{equation} 
\textbf{v}  =  \textbf{k}_p (\textbf{h}_{desired}-\textbf{h}) + \textbf{k}_d (\dot{\textbf{h}}_{desired} -\dot{\textbf{h}})
\end{equation}

The energy function can be plotted against time to check for convergence to zero. The energy of the now closed-loop system $E(t)$ in consideration is given by the sum of the potential and kinetic energies.

\begin{small}
\begin{eqnarray} 
E(t) &=& (\frac{1}{2}\,M\,\dot{x(t)}^2 +  \frac{1}{2}\,I_{yy}\,\dot{\theta(t)}^2) + (\frac{1}{2}\,k_{p1}\,(x_{desired}-x(t))^2 \nonumber\\ & &  + \frac{1}{2}\,k_{p2}\,(\theta_{desired}-\theta(t))^2) 
\end{eqnarray}
\end{small}

With appropriate controller gains and initial conditions, the energy equation was shown to converge to zero. From the expression $\ddot{\textbf{y}} = \textbf{v}$ it can be said that the system is stable for any positive damping coefficient.

The initial conditions have a significant influence on the response since the system is nonlinear. These include the x-position $x_0$, x-velocity $\dot{x}_0$, pitch angle $\theta_0$ and pitch rate $\dot{\theta}_0$. Through appropriate gain selection, the simulation was carried out with conservative limits for the thrust rates. The pitch rate was limited to $2\,rad/s$~\cite{sa2017dynamic}.

\begin{figure}[htpb]
\begin{center}
 
    \resizebox{42mm}{!} {\includegraphics{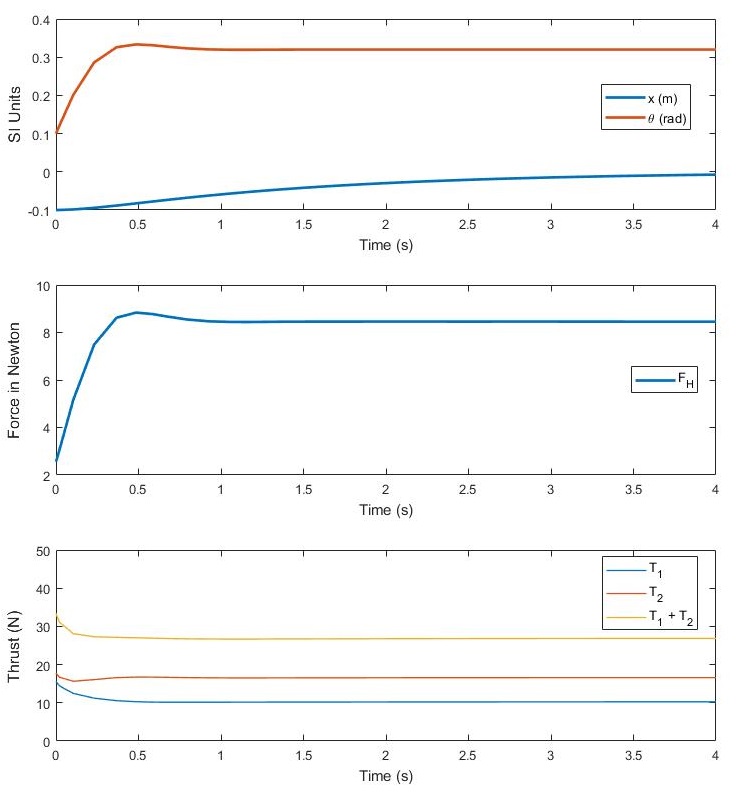}}
    \resizebox{42mm}{!} {\includegraphics{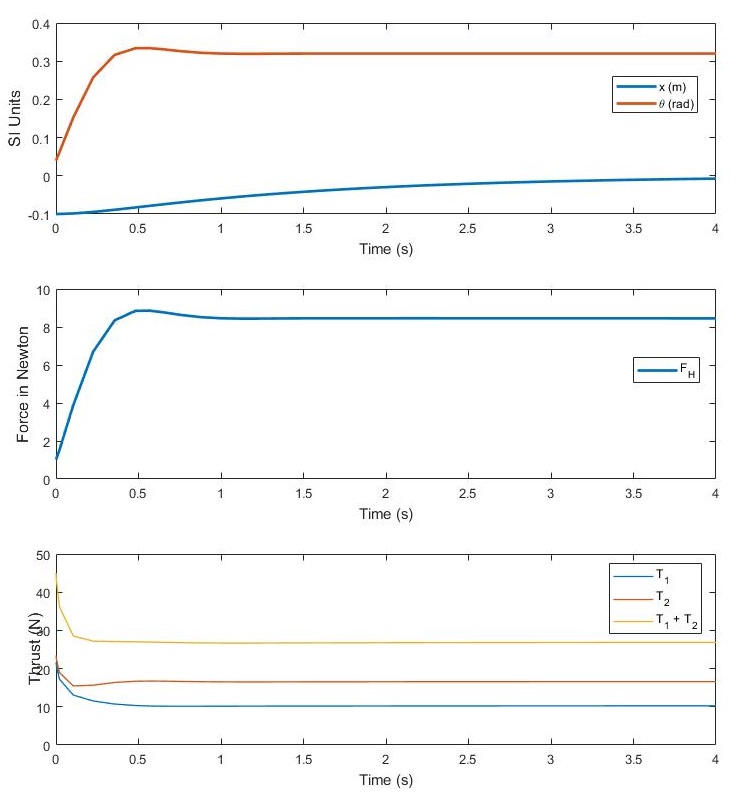}}
    \caption{(LEFT) Simulation results of closed-loop contact dynamics for acceptable initial conditions $x_0 = -0.1\,m$ , $\dot{x}_0 = 0\,m/s$ , $\theta_0 = 0.1\,rad$ , $\dot{\theta}_0 = 1\,rad/s$  regulating the state $x_{desired} = 0\,m$ and $F_{H,desired} = 8.5\,N $. (RIGHT) Simulation results of closed-loop contact dynamics for the initial conditions $x_0 = -0.1\,m$ , $\dot{x}_0 = 0\,m/s$ , $\theta_0 = 0.04\,rad$ , $\dot{\theta}_0 = 1\,rad/s$  regulating the state $x_{desired} = 0\,m$ and $F_{H,desired} = 8.5\,N $.}
  \label{fig:ideal}
  
  \end{center}
  
\end{figure}

Since the contact control mode is force-triggered, the initial value for force is about $2.5\,N.$ Figure (\ref{fig:ideal}) (left) shows that steady state is achieved within 0.5\,seconds for the force and the pitch angle. The settling time for the x-position is large but that is not relevant as the constantly running inverse kinematics engine of the manipulator will compensate for such errors. Additionally, the thrust values during the transient phase are well below their maximum value of 21\,N. Hence, this simulation result for this particular set of initial conditions is satisfactory.

There are certain cases that might occur during transition that may cause the controller to fail i.e. the required control inputs may exceed physical limits. These include cases in which the initial error from the reference values are large enough to cause undesired behaviour in the controller outputs $T_1$ and $T_2$.

\begin{itemize}

\item \textbf{Initial X-Position} $x_0$: This parameter has the least effect on the controller response as the errors associated will not be too large.

\item \textbf{Initial X-Velocity} $\dot{x_0}$: Ideally, this is zero as the velocity tends to be small when continuous contact is established. However, certain large positive values tend to reduce the required thrust considerably.

\item \textbf{Initial Pitch Angle} $\theta_0$: If the initial pitch angle reduces to smaller values (below 0.05\,rad), the required thrust for satisfactory reference increases rapidly as the contact dynamics are fast compared to the pitching response in figure (\ref{fig:ideal}) (right). A value too large would lead to dynamic instability as well as large force values. Since there exists a finite positive value of $\dot{\theta_0}$, a starting value less than the reference is desired. A value of 0.1\,rad is chosen since it results in an initial force close to the triggering value and a small overshoot.

\item \textbf{Initial Pitch Rate} $\dot{\theta_0}$: The worst and the most highly unlikely case is when this is a negative value. This can happen if large impacts create moments in the opposite direction.

\end{itemize}

According to the above discussion the control would transition smoothly from free-flight regime to contact regime when the following conditions are achieved:

\begin{enumerate}

\item \textbf{Detection of contact and a threshold force}: The end-effector must come into contact with the surface and must possess a threshold force value. To ensure smooth transition and elimination of chatter, a schmitt trigger approach is followed for this condition.

\item \textbf{Achieving the desired pose}: The pose that agrees with the above discussion as well as the manipulator base joint pose requirements for successful manipulation i.e. that the point on the surface to be applied  falls within the workspace and can be manipulated with the desired end-effector orientation.

\end{enumerate}

\section{IMPLEMENTATION}

The aerial platform used in this work is a DJI Matrice M100 quadrotor. It has a weight (with battery) of about 2400\,grams with a maximum take-off weight of 3600g. The lightweight nature of the manipulator is the result of clever design optimization as well as high strength-to-weight ratio materials like composites. The servomotor used for link 1 is the MKS DS1220 which is capable of providing 30.4\,kg.cm of torque. Link 2, which requires the largest amount of torque, used an MKS HV777 capable of a holding torque of 38\,kg.cm. For link 3, a faster and lighter servo was used -- Hitec HS77-BB.

\begin{figure}[ht]
	\centering
		\includegraphics[width=0.8\linewidth, height=6cm]{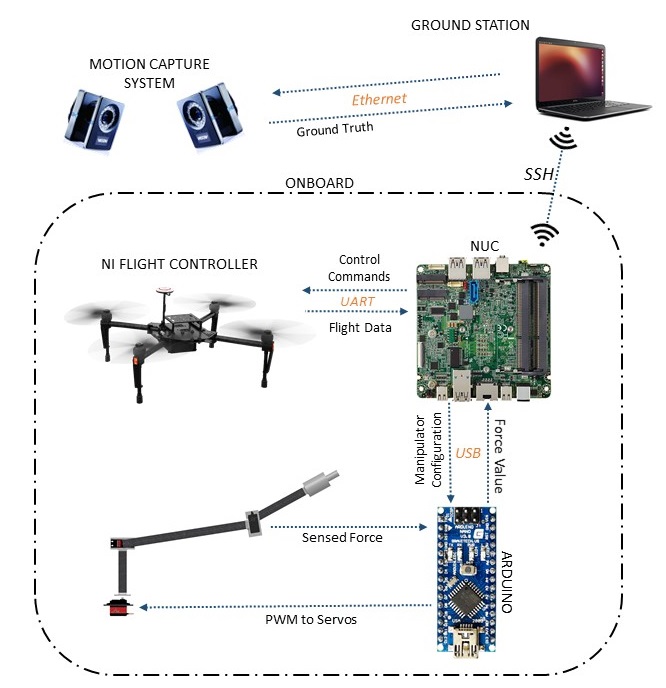}
		\caption[Hardware Signal Flow Diagram]{Hardware signal flow diagram showing the various hardware components, information flows and integration scheme for experimental testing using VICON. For operating in conditions with VICON absent, a visual-inertial localization scheme for the quadrotor along with visual servoing method for the manipulator can be used.}
	\label{fig:hdwr}
\end{figure}

The onboard processing unit was an Intel NUC i7. It runs a 3.5\,GHz  processor with 16\,GB of RAM.  An Arduino Nano microcontroller was used to interface between the servomotors for PWM control as well as to compute the force on the force sensor. It served as a device to receive serial commands from the Intel NUC containing desired manipulator configuration information and to send the value of force back to the NUC. The onboard computer ran ROS (Robotic Operating System) over a Linux OS. Nodes for high-level quadrotor control using MPC (Model Predictive Control), MCU interfacing, control transition and manipulator inverse kinematics were integrated for the inspection operation.

\section{CONCLUSIONS AND OUTLOOK}

In conclusion, this work demonstrated the design, controller simulation results and implementation scheme for NDT contact inspection using a three degree-of-freedom aerial manipulator.

One of the most important drawbacks of this work is the assumption of force regulation through pitch and 2-D position regulation alone. Thus, a more reliable state observer for force is needed.  Such state observers are included in techniques such as torque-based control and compliant control of manipulator interaction. Additionally, the  model,  although  intended  to  capture  the  low-level  nonlinear  dynamics,  did not consider the effect of the manipulator inertia on the system. From a design perspective, the serial manipulator in this work provides a large workspace and sufficient dexterity.  However, a parallel mechanism tends to be faster in kinematic reference tracking along the manipulated surface due to additional actuators. Another point of improvement is the overall design.  The off-center weights may be reduced by mounting all motors along the vertical axis of the quadrotor and use drive mechanisms such as pulley-belts or chain-sprockets.

\addtolength{\textheight}{-12cm}   





\bibliographystyle{IEEEtran}
\bibliography{./IEEEexample}

\end{document}